\documentclass[pre,superscriptaddress]{revtex4}
\usepackage{epsfig}
\usepackage{latexsym}
\usepackage{amsmath}
\usepackage{url}
\begin {document}
\title {Bio-linguistic transition and Baldwin effect in an evolutionary naming-game model}
\author{Adam Lipowski}
\affiliation{Faculty of Physics, Adam Mickiewicz University,
61-614 Pozna\'{n}, Poland}
\author{Dorota Lipowska}
\affiliation{Institute of Linguistics, Adam Mickiewicz University,
60-371 Pozna\'{n}, Poland}
\begin {abstract}
We examine an evolutionary naming-game model where communicating
agents are equipped with an evolutionarily selected learning
ability. Such a coupling of biological and linguistic ingredients
results in an abrupt transition: upon a small change of a model
control parameter a poorly communicating group of linguistically
unskilled agents transforms into almost perfectly communicating
group with large learning abilities. When learning ability is kept
fixed, the transition appears to be continuous. Genetic imprinting
of the learning abilities proceeds via Baldwin effect: initially
unskilled communicating agents learn a language and that creates a
niche in which there is an evolutionary pressure for the increase
of learning ability.Our model suggests that when linguistic (or cultural) processes became intensive enough, a  transition took place where both linguistic performance and biological endowment of our species experienced an abrupt change that perhaps triggered the rapid expansion of human civilization.
\end{abstract}
\maketitle
\section{Introduction}
Recently, the emergence and evolution of language attracts a growing
interest. In this interdisciplinary field problems like
differentiation of languages, development of the speech apparatus
or  formation of linguistically connected social groups require
joint efforts of many specialists  such as
linguists, neuroscientists, or anthropologists. However, there are also
more general questions concerning this almost exclusively human
trait. Why do we use words and then combine them into sentences?
Why all languages have grammar? To what extent is our brain
adapted for acquisition of language? Can learning direct the
evolution? This sample questions justify an increasing
involvement of researchers also from other disciplines such as  artificial intelligence, computer sciences, evolutionary biology or physics~\cite{NOWAK}.

Computer modeling is a frequently used tool in the studies of language
evolution. In this technique two main approaches can be
distinguished. In the first one, known as an iterated learning
model, one is mainly concerned with the transmission of language
between successive generations of agents~\cite{KIRBY2001,BRIGHTON2002}. The important issue that the iterated learning model has successfully addressed is the transition from holistic to compositional language. However, since the number of communicating agents is typically very small, the problem of the emergence of linguistic coherence must be neglected in this approach. To tackle this problem Steels introduced a naming game model~\cite{STEELS1995}. In this approach one examines a population of agents trying to establish a common vocabulary for a certain number of objects present in their environment. The change of generations is not required in the naming game-model since the emergence of a common vocabulary is a consequence of the communication processes between agents.

It seems that the iterated learning model and the naming-game model are at two extremes: the first one emphasizes the generational turnover while the latter concentrates on the single-generation (cultural) interactions. Since in the language evolution both aspects are present, it is desirable to examine models that combine evolutionary and cultural processes. In the present paper we introduce such a model. Agents in our model try to establish a common vocabulary like in the naming-game model, but in addition they can breed, mutate, and die. Moreover, they are equipped with an evolutionary trait: learning ability.
As a result evolutionary and cultural (learning from peers) processes mutually influence each other. When communication between agents is sufficiently frequent, cultural processes create a favourable niche in which a larger learning ability becomes advantageous. But gradually increasing learning abilities in turn speed up the cultural processes. As a result the model undergoes an abrupt bio-linguistic transition. One can speculate that the proposed model suggests that linguistic and biological processes at a certain point of human history after crossing a certain threshold started to have a strong influence on each other and that resulted in an explosive development of our species. That learning in our model modifies the fitness landscape of a given agent and facilitates the genetic accommodation of learning ability is  actually a manifestation of the much debated Baldwin effect~\cite{YAMAUCHI2004,TURNEY1996}.
\section{Model}
In our model we consider a set of agents located at sites of the square lattice of the linear size $L$. Agents are trying to establish a common vocabulary on a single object present in their environment. An assumption that agents communicate only on a single object does not seem to restrict the generality of our considerations and was already used in some other studies of naming-game~\cite{BARONCHELLI2006,ASTA2006} or language-change~\cite{NETTLE1999} models.
A randomly selected agent takes the role of a speaker that communicates a word chosen from its inventory to a hearer that is randomly selected among nearest neighbours of the speaker. The hearer tries to recognize the  communicated word, namely it checks whether it has it in its inventory. A positive or negative result translates into communicative success or failure, respectively. In some versions of the naming-game model~\cite{BARONCHELLI2006,ASTA2006} success means that both agents retain in their inventories only the chosen word while in the case of failure the hearer adds the communicated word to its inventory.

To implement the learning ability we modified this rule and assigned weights $w_i$ ($w_i>0$) to each $i$-th word in the inventory. The speaker selects then the $i$-th word  with the probability $w_i/\sum_j w_j$ where summation is over all words in its inventory (if its inventory is empty, it creates a word randomly). If the hearer has the word in its inventory, it is recognized.
In addition, each agent $k$ is characterized by its learning ability $l_k$ ($0<l_k<1$) that is used to modify weights. Namely, in the case of success both speaker and hearer increase the weights of the communicated word by learning abilities of the speaker and hearer, respectively. In the case of failure the speaker subtracts its learning ability from the weight of the communicated word. If after such a subtraction a weight becomes negative the corresponding word is removed from the repository. The hearer in the case of failure, i.e., when it does not have the word in its inventory, adds the communicated word to its inventory with a unit weight.

In addition to communication, agents in our model evolve according
to the population dynamics: they can breed, mutate, and eventually
die. To specify intensity of these processes we introduce the
communication probability $p$. With probability $p$ the chosen
agent becomes a speaker and with probability $1-p$ we will attempt
a population update. During such a move the agent dies with the
probability $1-p_{\rm surv}$, where $p_{\rm surv}=\exp (-at)[
1-\exp(-b\sum_j w_j/\langle w\rangle)]$ and $a\sim 0.05$ and $b=5$
are certain parameters whose role is to ensure a certain speed of
population turnover. Moreover, $t$ is the age of an agent and
$\langle w\rangle$ is the average (over agents) sum of weights.
Such a formula takes into account both its linguistic performance
(the bigger $\sum_j w_j$ the larger $p_{\rm surv}$) and its age.
If the agent survives (it happens with the probability $p_{\rm
surv}$) it breeds, provided that there is an empty site on one of
the neighbouring sites. The offspring typically inherits parent's
learning ability and the word from its inventory that has the
highest weight. In the offspring inventory the weight assigned
initially  to this word equals one. With the small probability
$p_{\rm mut}$ a mutation takes place and the learning ability of
an offspring is selected randomly anew. With the same probability
an independent check  is  made whether to mutate the inherited
word. A diagram illustrating the dynamics of our model is given in
the Appendix~\cite{applet}. Let us also notice that the behaviour
of our model, that is described below, is to some extent robust
with respect to some modifications of its rules. For example,
qualitatively the same behaviour is observed for modified
parameters $a$ and $b$, different form of the survival probability
$p_{\rm surv}$ (provided it is a decreasing function of $t$ and an
increasing function of $\sum_j w_j$), or different breeding and/or
mutation rules.

\section{Results}
To examine the properties of the model we used numerical simulations. Most of the results are obtained for $L=40$ and $p_{\rm mut}=0.001$ but simulations for $L=60$ and $p_{\rm mut}=0.01$ lead to the similar behaviour. Simulations start from all sites occupied by agents and having a single, randomly chosen for each agent word in their inventories with a unit weight. The learning ability of each agent is also chosen randomly.
\subsection{Bio-linguistic transition}
\begin{figure}
\vspace{2.5cm} \centerline{ \epsfxsize=13cm \epsfbox{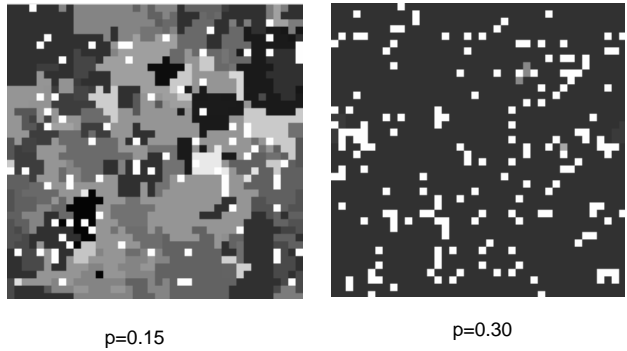} }
\vspace{-6cm} \caption{Exemplary configurations of the evolutinary
naming-game model with $L=40$ and  $p_{\rm mut}=0.001$. In the
small-$p$ phase (left panel) communications are infrequent and
agents using the same language form only small clusters. In this
phase communicative success rate $s$ and learning ability are
small (see Fig.~\ref{steady}). In the large-$p$ phase (right
panel) frequent communications result in the emergence of the
common language.}\label{conf}
\end{figure}

\begin{figure}
\vspace{1cm} \centerline{ \epsfxsize=9cm \epsfbox{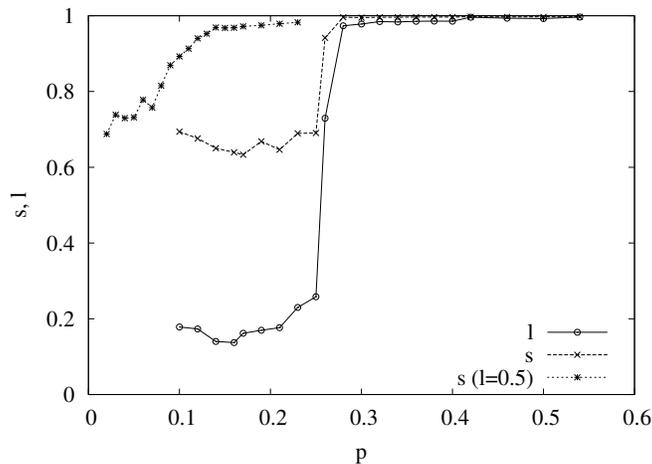} }
\vspace{5mm} \caption{Success rate $s$ and learning ability $l$ as
a function of communication probability $p$. Calculations were made
for system size $L=40$ and mutation probability $p_{\rm mut}=0.001$.
Simulation time was typically equal to $10^5$ steps with $3\cdot 10^4$ steps discarded for relaxation. A step is defined as a single, on average, update of each site.
The data in the left upper corner are obtained for simulations with learning ability kept fixed.}
\label{steady}
\end{figure}

An important parameter of the model is the communication probability $p$ that specifies intensity of communication attempts in comparison with populational changes. In general, for small $p$ the model remains in the phase of linguistic disorder with only small clusters of agents using the same language. We define the language of an agent as the  largest-weight word in its inventory. Such a definition means that agents using the same language usually (but not always) use a recognizable word and it ensures a relatively large rate of communication successes of such agents. A typical distribution of languages in this disordered small-$p$ phase is shown in the left panel of Fig.~\ref{conf} where agents using the same language  are drawn with the same shade of grey. Upon increasing the communication probability $p$ the clusters of agents only slightly increase, but after reaching a certain threshold an abrupt transition takes place and the model enters the phase of linguistic coherence with almost all agents belonging to the same cluster (Fig.~\ref{conf}, right panel).
To examine the nature of this transition we measured the communication success rate $s$ defined as an average over agents and simulation time of the fraction of successes with respect to all communication attempts. Moreover, we measured the average learning ability $l$. The measured values of $s$ and $l$ as a function of $p$ are shown in Fig.~\ref{steady}. One can notice that the abrupt transition around $p=0.25$ manifests not only through the jump of the communicative properties ($s$) but also through the jump of the biological endowment (jump of $l$).
Such a coincidence is by no means obvious and in principle one can imagine these two transitions being separated. Before examining the mechanism responsible for such an agreement let us mention about the behaviour of the model with the learning ability kept fixed during entire simulations. In this case there is also a phase transition between disordered and linguistically coherent phases but this time the transition is much smoother (Fig.~\ref{steady}).
Large fluctuations of the success rate $s$ in the vicinity of the transition and absence of the jump suggest that this might be a continuous transition.
In the last section we will return to this point. Let us also notice that sudden transitions in linguistic models were also reported in some other models~\cite{NOWAK,STAUFFER2008}.
\subsection{Baldwin effect}
The agreement of the transitions as seen in Fig.~\ref{steady} shows that communicative and biological ingredients in our model strongly influence each other and that leads to the single and abrupt transition. In our model successful communication requires learning. A new-born agent communicating with some mature agents who already  worked out a certain (common in this group) language will increase the weight of a corresponding word. As a result, in its future communications the agent will use mainly this word. In what way such a learning might get coupled with evolutionary traits? The explanation of this phenomenon is known as a Baldwin effect. Although at first sight it looks like a discredited Lamarckian phenomenon, Baldwin effect is actually purely Darwinian\cite{HINTON1987,YAMAUCHI2004}. There are usually some benefits related with the task a given species has to learn and there is a cost of learning this task. One can argue that in such a case there is some kind of evolutionary pressure that favours individuals for which the benefit is larger or the cost is smaller. Then, evolution will lead to the formation of species where the learned behaviour becomes an innate ability. It should be emphasized that the acquired characteristics are not inherited. What is inherited is the ability to acquire the characteristics (the ability to learn)~\cite{TURNEY1996}. In the context of language evolution the importance of the Baldwin effect was suggested by Pinker and Bloom~\cite{PINKER1990}. Perhaps this effect is also at least partially responsible for the formation of the Language Acquisition Device - the hypothetical structure in our brain whose existence was suggested by Chomsky. However, many details concerning the role of the Baldwin effect in the evolution of language remain unclear~\cite{MUNROE2002}.

\begin{figure}
\vspace{1cm} \centerline{ \epsfxsize=9cm \epsfbox{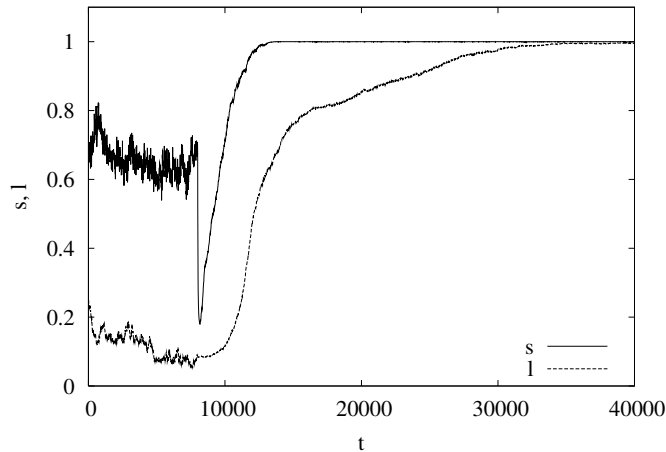} }
\vspace{5mm} \caption{Time dependence of  the success rate $s$ and
learning ability $l$. Initially, the system evolves with
communication probability $p=0.1$ that at time $t=8000$ is increased
to $p=0.98$.
Such a large value of $p$ means that agents are mainly communicating and that enables us to notice that fast cultural processes (increase of the success rate, upper line) are followed by slow evolutionary changes (increase of the learning ability).
A transient decrease of the learning ability just after increase of $p$ is an unimportant numerical artifact, that disappears when the jump of $p$ is replaced by a continuous change.
Calculations were made for system size $L=40$ and
mutation probability $p_{\rm mut}=0.001$.} \label{time}
\end{figure}

In our model the Baldwin effect is also at work. Let us consider a population of agents  with the communication probability $p$ below the threshold value ($p=p_c\approx 0.25$). In  such a case learning ability remains at a rather low level (since clusters of agents using the same language are small, it does not pay off to be good at learning the language of your neighbours). Now, let us increase the value of $p$ above the threshold value (Fig.~\ref{time}).
More frequent communication changes the behaviour dramatically. Apparently, clusters of agents using the same language are now sufficiently large and it pays off to have a large learning ability because that  increases the success rate and thus the survival probability $p_{\rm surv}$. Let us notice that $p_{\rm surv}$ of an agent depends on its linguistic performance ($\sum_j w_j$) rather than its learning ability. Thus clusters of agents of good linguistic performance (learned behaviour) can be considered as niches that direct the evolution by favouring agents with large learning abilities, which is precisely the Baldwin effect. It should be noticed that linguistic interactions between agents (whose rate is set by the probability $p$) are typically much faster than evolutionary changes (set by $p_{\rm mut}$). To observe such a difference in our simulations shown in Fig.~\ref{time} we  increased $p$ up to the value 0.98. And indeed, the linguistic changes (success rate), that might be considered as niches-forming processes, are ahead of the evolutionary adaptations (learning ability).

As a result of a positive feedback (large learning ability enhances communication that enlarges clusters  that favours even more the increased learning ability) a discontinuous transition takes place both with respect to the success rate and learning ability (Fig.~\ref{steady}). An interesting question is whether such a behaviour is of any relevance in the context of human evolution. It is obvious that development of language, which probably took place somewhere around $10^5$ years ago, was accompanied by important anatomical changes such as fixation of the so-called speech gene (FOXP2), descended larynx or enlargement of brain~\cite{HOLDEN2004}. Linguistic and other cultural interactions that were already emerging in early hominid populations were certainly shaping the fitness landscape and that could direct the evolution of our ancestors via the Baldwin effect. Our model predicts that when intensity of linguistic (or cultural) processes was large enough, a  transition took place where both linguistic performance and biological endowment of our species experienced an abrupt change that perhaps lead to the rapid expansion of human civilization. But further research would be needed to claim that such a transition did take place and explain it within the framework suggested in our paper.
\section{Conclusions}
In the present paper we examined an evolutionary naming-game model.
Simulations show that coupling of linguistic and evolutionary ingredients produces a discontinuous transition and learning can direct the evolution towards better linguistic abilities (Baldwin effect).
The present model computationally is not very demanding. It seems to be  possible to consider agents talking on more than one object, or to examine statistical properties of simulated languages such as for example, distributions of their lifetimes or of the number of users. One can also study effects like diffusion of languages, the role of geographical barriers, or formation of language families.
There is already an extensive literature documenting linguistic data as well as various computational approaches modeling for example competition between already existing natural languages ~\cite{ABRAMS2003,STAUFFER2008,PAULO2007}.
The dynamics of the present model, that is based on an act of elementary communication, offers perhaps more natural description of dynamics of languages than some other approaches that often use some kind of coarse-grained dynamics.

There are also more physical aspects of the proposed model that might be worth further studies. As we have already mentioned, when the learning ability is kept fixed, the transition between disordered and linguistically coherent phases seems to be continuous. On the other hand, such a transition resembles the symmetry breaking transition in the $q$-state Potts model, where at sufficiently low temperature the model collapses on one of the $q$ ground states. However, in the two dimensional case and for large $q$ ($q$ in our case corresponds to the number of all languages used by agents) such a transition is known to be discontinuous. Of course the dynamics of our model is much different from Glauber or Metropolis dynamics  that reproduce the equilibrium Potts model, but very often such differences are irrelevant as long as, for example, the symmetry of the model is preserved (which is the case for our model). Another possibility that would explain a continuous nature of the transition in our case might be a different nature of (effective) domain walls between clusters. In our model these domain walls in some cases might be much softer and that would shift the behaviour of our model toward models with continuous-like symmetry (as e.g., XY model). To clarify this issue further work is, however, needed.\\

{\bf Acknowledgments:} We gratefully acknowledge access to the computing
facilities at Pozna\'n Supercomputing and Networking Center.
\newpage
\begin{minipage}{18cm}
{
{\bf Appendix}\\
\par
The following diagram illustrates an elementary step of the dynamics.The change of inventories in the case of {\it success} or {\it failure} is described in the text.\\
\centerline{ \epsfxsize=17cm \epsfbox{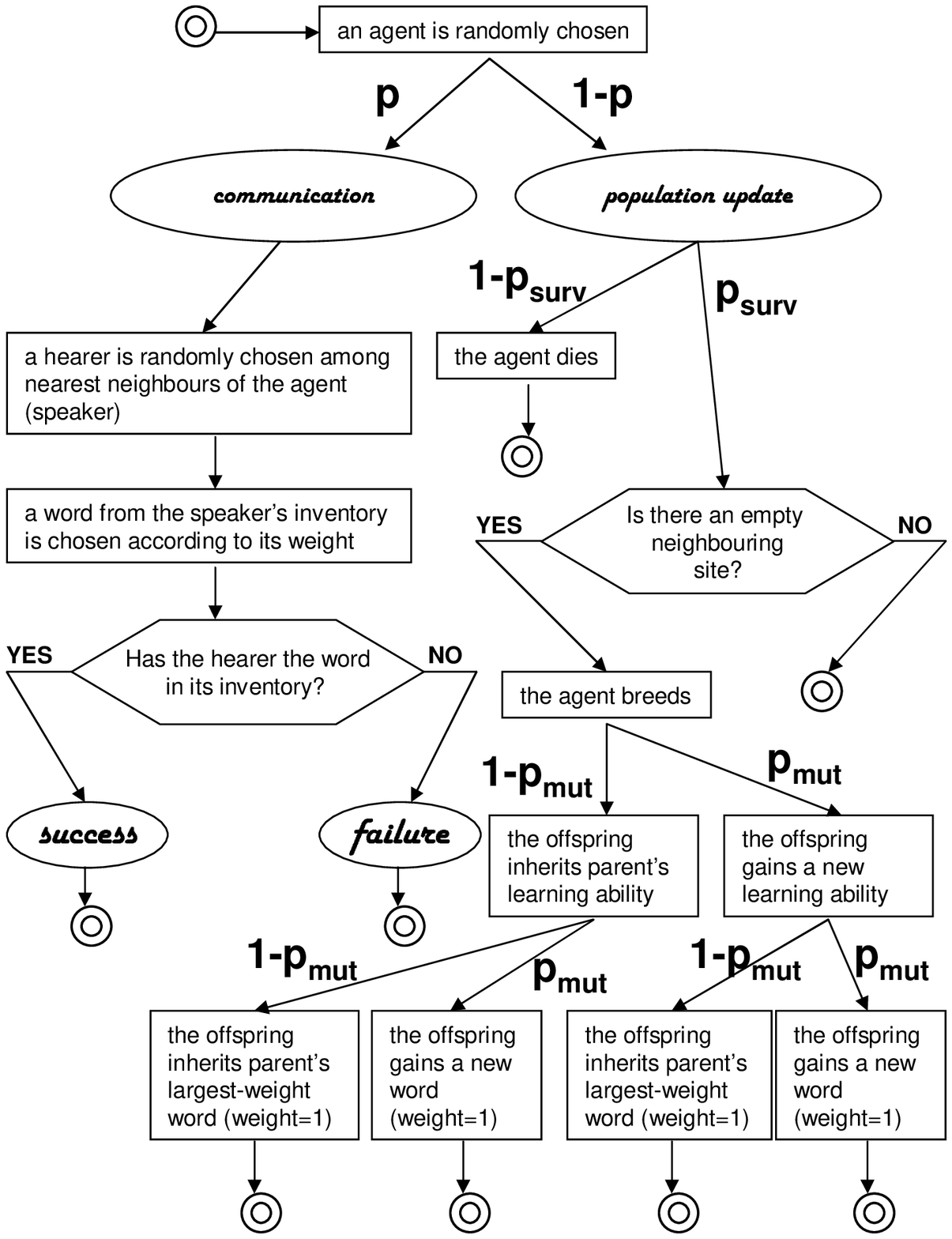} }
}
\end{minipage}

\end {document}